\documentclass{article}
\usepackage{amsmath}

\usepackage[preprint]{neurips_2024}

\newcommand{\OURS}{DT-NVS}

\usepackage[utf8]{inputenc} 
\usepackage[T1]{fontenc}    
\usepackage{hyperref}       
\usepackage{url}            
\usepackage{booktabs}       
\usepackage{amsfonts}       
\usepackage{nicefrac}       
\usepackage{microtype}      
\usepackage{xcolor}         
\usepackage{graphicx} 
\usepackage{wrapfig}

\title{\OURS{}: Diffusion Transformers for\\ Novel View Synthesis}

\author{
Wonbong Jang\textsuperscript{*} \quad Jonathan Tremblay\textsuperscript{†} \quad
Lourdes Agapito\textsuperscript{*}  \\[0.1cm]
\textsuperscript{*}UCL \qquad \textsuperscript{†}NVIDIA \\[-0.1cm]
{\tt\small \{ucabwja,l.agapito\}@ucl.ac.uk} \quad
{\tt\small  jtremblay@nvidia.com}
}

\begin{document}

\maketitle

\begin{abstract}

Generating novel views of a natural scene, e.g., every-day scenes both indoors and outdoors, from a single view is an under-explored problem, 
even though it is an organic extension to the object-centric novel view synthesis. Existing diffusion-based approaches focus rather on small camera movements in real scenes or only consider unnatural object-centric scenes, limiting their potential applications in real-world settings. 
In this paper we move away from these constrained regimes and propose a 3D diffusion model trained with image-only losses on a large-scale dataset of real-world, multi-category, unaligned, and casually acquired videos of everyday scenes. We propose \OURS{}, a 3D-aware diffusion model for generalized novel view synthesis that exploits a transformer-based architecture backbone. We make significant contributions to transformer and self-attention architectures to translate images to 3d representations, and novel camera conditioning strategies to allow training on real-world unaligned datasets. In addition, we introduce a novel training paradigm swapping the role of reference frame between the conditioning image and the sampled noisy input. We evaluate our approach on the 3D task of generalized novel view synthesis from a single input image and show improvements over state-of-the-art 3D aware diffusion models and deterministic approaches, while generating diverse outputs.

\end{abstract}

\section{Introduction}


Diffusion models have emerged as a powerful methodology for high-quality 2D image and video generation from multimodal inputs. However, training diffusion models to learn 3D representations for truly 3D-aware generation has not been straightforward. 
%
The unique challenge comes from their reliance on large amounts of ground-truth training data which, in the case of 3D scenes, is scarce and costly to acquire. 
Meanwhile, recent advances in 3D geometry and appearance acquisition from images have resulted in powerful methods such as neural radiance fields (NeRFs)~\cite{mildenhall2021nerf}, InstantNGP~\cite{muller2022instant} or Gaussian Splatting~\cite{kerbl3Dgaussians}, which can learn 3D implicit scene representations for high-quality new view synthesis from 2D images only, without the need for 3D ground truth. However, these methods are scene-specific, they require a large number of carefully acquired input views with corresponding camera poses, and need costly test-time optimization.

In this paper we focus on the challenging problem of generating novel views of general scenes, from a single input image and using only 2D losses. Numerous approaches have aimed to generalize NeRF to model multiple scenes; 
some explored using global latents~\cite{jang2021codenerf}, 
back-projecting features from pre-trained networks~\cite{yu2021pixelnerf}, 
or applying generative models such as GANs~\cite{eg3dChan2021}. 
More recently, we have witnessed attempts to train 3D diffusion models for this task with 2D-only supervision using implicit representations~\cite{anciukevivcius2023renderdiffusion,szymanowicz2023viewset, karnewar2023holodiffusion}. 
However, they require canonicalized datasets with 3D-aligned scenes, assume single-category, object-centric scenes and simplified camera models, or they rely on synthetic data. 
The recent emergence of large-scale multi-view datasets of casually-captured videos of hundreds of object categories such as Co3D~\cite{reizenstein2021common} or MVImgNet~\cite{yu2023mvimgnet} has opened the door for transformer-only architectures also to be proposed for novel-view synthesis~\cite{hong2023lrm,jang2023nvist}. 
While these methods show promising results, they are deterministic at heart, while the task of novel view synthesis is clearly a generative one.




%
%
We propose \OURS{}, a novel view synthesis diffusion model that exploits a transformer-based backbone architecture to predict a radiance field from a single reference image. 
%
Unlike many diffusion-based solutions to novel view synthesis, our approach is not limited to masked, object-centric scenes, or 3D aligned scenes, and can be applied to any real-world captures.
As such, we propose new self-attention architectures along with novel camera viewpoint conditioning strategies and we introduce a novel training paradigm that switches between sampled noise and reference images to avoid trivial or degraded solutions.
We leverage MVImgNet~\cite{yu2023mvimgnet}, a very large-scale dataset of multi-view videos of indoor and outdoor scenes of hundreds of categories of everyday objects. 
Our model quantitatively achieves better FID scores than deterministic transformer-based architectures such as NViST~\cite{jang2023nvist} and also outperforms non-transformer based diffusion models ~\cite{anciukevivcius2023renderdiffusion,szymanowicz2023viewset,anciukevicius2024denoising}, 
on both real-world (42\% increase) and synthetic (212\% increase) datasets. 
We also provide a detailed ablation study to justify our design choices.

\begin{figure}[tb]
    \centering
    \includegraphics[width=\textwidth]{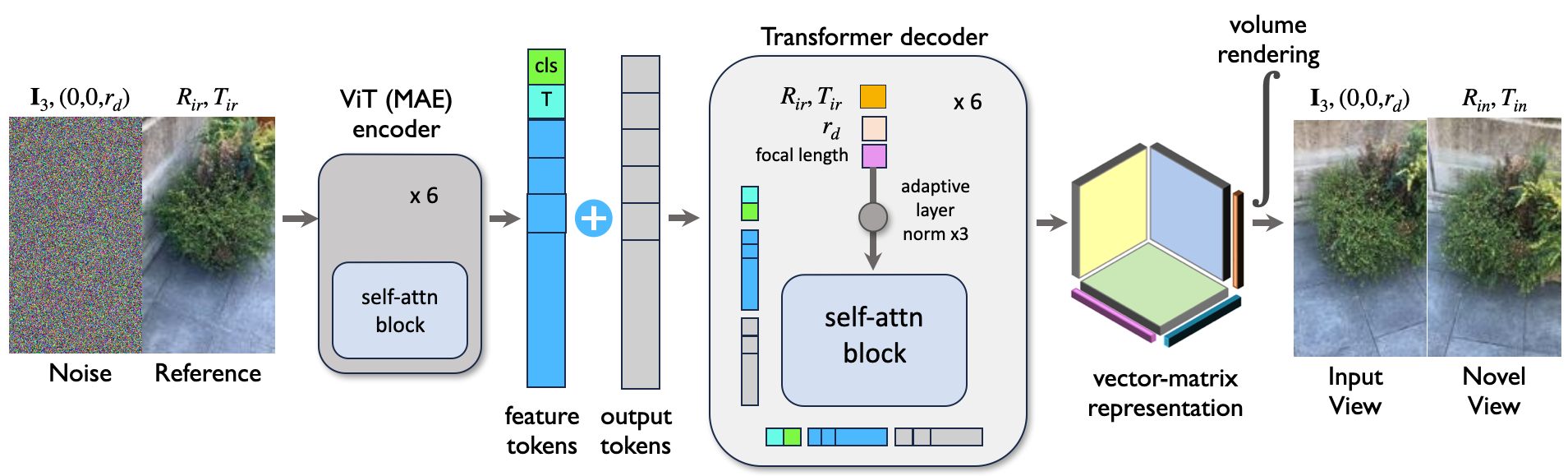}
    \caption{
    \textbf{Architecture: }
    \OURS{} is a 3D-aware diffusion model that takes noise $z^i_t$ at $I_3$ and a reference image $x^r$ at reference viewpoints $R_{ir}, T_{ir}$ as input and learns to denoise from $c^i$ and do novel view synthesis on $R_{in}, T_{in}$.
    The encoder processes noise $z$ and the reference image $x^r$ separately, and generates feature tokens for each.
    Our decoder concatenates both feature tokens with its output tokens, then applies self-attention with conditioning tokens on their respective camera parameters (input view for output tokens and feature tokens from $z_i^t$, and reference viewpoints on features tokens from $x^r$) 
    We reshape the output tokens into vector-matrix representation, and perform volume rendering. 
    During the training, we supervise the model with denoising loss at input view $I_3$ and photometric loss on novel view. 
    For inference, we use the predicted output at input view $\smash{\hat{x^i}}$ and denoise according to diffusion step $t$.}
    \label{fig:overall}
\end{figure}

\section{Related Work}


\textbf{Diffusion Models:} A diffusion model is a generative approach similar to GANs, first proposed by Sohl-Dickstein \textit{et al.}~\cite{sohl-dickstein2015deep}, and gained popularity with DDPM~\cite{ho2020denoising}. 
Various techniques have been proposed to enhance the quality of diffusion model outcomes, including Cosine Schedule~\cite{dhariwal2021diffusion}, v-parameterization~\cite{salimans2022progressive}, and classifier-free guidance~\cite{ho2022classifier}. 
DDIM~\cite{song2020denoising} offers a more flexible and efficient sampling process, enabling faster generation of high-quality images with fewer steps.
Hang \textit{et al.}~\cite{hang2023efficient} proposed a minimum signal-to-noise ratio (Min-SNR) strategy, showing that the model converges faster by using SNR as a weight to the loss function.
DiT~\cite{peebles2023scalable} scales the Diffusion model with Transformers by conditioning diffusion steps with Adaptive Layer Normalization(AdaLN)-Zero~\cite{xu2019understanding}.

\textbf{Neural Radiance Fields: } Neural 3D implicit representations were initially proposed in  SRN~\cite{sitzmann2019scene} and later used in DVR~\cite{niemeyer2020differentiable}, to train 3D-aware representations without 3D ground-truth.
NeRF~\cite{mildenhall2021nerf} revolutionized novel view synthesis from collections of posed images.
To accelerate NeRF training, grid-based representations have been proposed. ~\cite{yu2021plenoxels, muller2022instant,fridovich2023k, tilted2023} EG3D applied triplanes by projecting features into three planes in the context of 3D-aware GANs and TensoRF~\cite{chen2022tensorf} proposed the vector-matrix representation.
Efforts to generalize NeRF include conditioning on global latent vectors~\cite{Gafni_2021_CVPR, jang2021codenerf, muller2022autorf,rebain2022lolnerf,athar2022rignerf,hong2022headnerf}, associating 2D feature views with target views~\cite{yu2021pixelnerf, wang2021ibrnet, chen2021mvsnerf,reizenstein2021common, Henzler_2021_CVPR, trevithick2021grf,chen2023matchnerf, irshad2023neo360}, or supervising NeRF with GAN losses~\cite{chanmonteiro2020pi-GAN, eg3dChan2021, schwarz2020graf,le2022stylemorph,cai2022pix2nerf}. 

\textbf{3D-Aware Diffusion Models: } Extending  diffusion models to be 3D-aware has proved challenging due to the lack of large 3D ground truth datasets.
Previous approaches fine-tuned a pre-trained latent diffusion model~\cite{rombach2022high} on camera parameters~\cite{liu2023zero, melas2023realfusion, shi2023mvdream, tang2023dreamgaussian, wang2023prolificdreamer, zeronvs, watson2022novel}
Other approaches~\cite{poole2022dreamfusion, tang2023dreamgaussian, wu2023reconfusion, wang2023morpheus} regularize NeRF using the pre-trained diffusion models or utilize the pre-trained diffusion models to generate images.
Similar to latent diffusion, several approaches first train the 3D-aware GAN and apply the diffusion process on latent vectors~\cite{Schwarz2024ICLR, kim2023nfldm}.
Instead of 2-stage training, there are approaches other methods obtain intermediate features and use them to denoise from the target view ~\cite{chan2023genvs, gu2023nerfdiff, tewari2023forwarddiffusion}.
Another line of research involves learning diffusion models in 3D by rendering in 2D or backprojecting 2D features into 3D~\cite{karnewar2023holodiffusion, anciukevicius2024denoising, anciukevivcius2023renderdiffusion, szymanowicz2023viewset}, which usually assume the aligned scenes.

\textbf{Transformers for 3D Tasks: } Vision Transformer (ViT)~\cite{dosovitskiy2020image} has been successful in the field of computer vision, and 
Masked Autoencoder(MAE)~\cite{he2022masked} learns visual features in self-supervised learning. 
For 3D, geometry-free methods based on transformer have been explored in ~\cite{kulhanek2022viewformer, sajjadi2022scene, Miyato2024GTA}.
To build 3D implicit representation from Transformer, GINA3D~\cite{shen2023gina} and VQ3D~\cite{sargent2023vq3d} employ the adversarial loss.
Other approaches deal with point clouds~\cite{jun2023shap, wu2023multiview}.
LRM~\cite{hong2023lrm} and NViST~\cite{jang2023nvist} apply the Transformer for 3D implicit representation in a deterministic way, and DMV~\cite{xu2023dmv3d} extends LRM into a diffusion, but both LRM and DMV do not model the background.
\OURS{} is the first method to apply the Transformer with a 3D-aware diffusion model, directly learning from real-world scenes.

\section{Background}

For each scene $\mathcal{S}$, we have RGB renderings $x^i$ from camera poses $c^i$ and focal length $f$ (which we assume constant).  
A diffusion model generates an image using a series of iterative steps, from noisy input to generated content. 
During this process, the output of the model is fed back as input recursively until convergence. 
To train such model, a forward process involves gradually adding noise to an image over several steps until it becomes nearly pure noise. 
Then, during the reverse process, the method aims to reverse the noise addition, 
transforming the noisy image back into the original image.

Formally, during the forward process, 
noise is added $\epsilon\sim\mathcal{N}(0,I)$ to $x^i$ to form $z_t^i = \alpha_t x^i + \sigma_t \epsilon$ at each diffusion step $t \in [1,T]$. 
We assume a variance-preserving scenario, where $\alpha_t^2+\sigma_t^2=1$, and follow the cosine schedule proposed in~\cite{dhariwal2021diffusion}, with $\alpha_t = \cos(\pi t / 2T)$ and $\sigma_t = \sin(\pi t / 2T)$. 
With the assumptions of a Gaussian distribution and Markov chain, we can represent the marginal as $q(z_t^i | x^i)=\mathcal{N}(z_t^i|\alpha_t x^i, \sigma_t I)$. 
Given $t > s$, $q(z_t^i | z_s^i) = \mathcal{N}(z_t^i|\alpha_{t|s} z_s^i, \sigma_{t|s}I)$, where $\alpha_{t|s} = \alpha_t/\alpha_s$ and $\sigma_{t|s} = \sigma_t^2 - \alpha_{t|s}^2 \sigma_s^2$.

In the reverse process, $q(z_s^i|z_t^i)$ is intractable; however, there exists a closed form solution when we condition on $x^i$ as $q(z_s^i|z_t^i, x^i)=\mathcal{N}(z_t^i|\mu_{t\rightarrow s}, \sigma_{t\rightarrow s}^2 I)$, where $\mu_{t\rightarrow s} = \alpha_{t|s} \sigma_s^2 / \sigma_t^2 z_t^i + \alpha_s \sigma_{t|s}^2 / \sigma_t^2 x^i$ and $\sigma_{t\rightarrow s}^2 = \sigma_{t|s}^2 \sigma_s^2 / \sigma_t^2$. 
Applying the result from~\cite{ho2020denoising}  $s\rightarrow t$, we can represent $p(z_s^i|z_t^i) = q(z_s^i|z_t^i, x^i = \hat{x}^i)$, using $\hat{x}^i$. 
Therefore, by estimating $\hat{x}^i$, we can approximate the posterior $q(z_s^i|z_t^i, x = \hat{x}^i)$. 

Since our diffusion model needs to perform novel-view synthesis during training, our diffusion loss cannot be based on the noise ($\epsilon$-parameterization) as  DDPM~\cite{ho2020denoising}, but rather on the ground truth ($x$-parameterization). To improve convergence we take inspiration from~\cite{salimans2022progressive, ho2022imagen} who proposed the $v$-parameterization, a re-weighting of the $\epsilon$-parameterization: $v_t^i=\alpha_t\hat{\epsilon} - \sigma_t\hat{x}^i$. More concretely, we follow Hang \textit{et al.}~\cite{hang2023efficient} who showed that all three parameterizations $x$, $v$ and $\epsilon$ are equivalent by reweighting the $x$ parameterization with signal-to-noise ratio values (SNR) $\alpha_t^2 / \sigma_t^2$.

\section{Methodology}

\subsection{Diffusion in 3D implicit representation}

In 3D implicit representations, $\mathcal{S}$ is not directly observable since we only have access to a reference view. 
Therefore, we add noise $\epsilon$ to scene images $x^i$ 
and train the model $\mathcal{F}_\theta$ to denoise them from the camera viewpoint $c^i$.
To be 3D-aware, the model should not only estimate the image $x^i$ but also predict all of the possible images, $x^n$, in the scenes 
from their viewpoints, $c^n$.
Here, we assume that we do not observe the noise level from $c^n$, so the model estimates $\hat{x}^n$, which means that the model predicts the ground-truth. 
For each diffusion step $t$, we multiply the weight $w(t)$=min (SNR(t)+1,5), so that it is similar to $v$-parameterization.
We can write the denoising loss $\mathcal{L}_{denoising}$ as below.

\begin{equation}
    \mathcal{L}_{denoising} = \mathbb{E}_{\theta,t,x^r}[w(t)(\hat{x}^i - x^i)^2]
    \label{eq:denoising}
\end{equation}

We get $\hat{x}^n$ by rendering the estimated scene $\hat{\mathcal{S}}$ generated from the model $\mathcal{F}_\theta$ which takes the reference image $x^r$, noisy image $z_t^i$, and rays from the camera viewpoints $c^i$ and $c^r$.

\begin{wrapfigure}{r}{0.5\textwidth}
    \centering
    \includegraphics[width=0.5\textwidth]{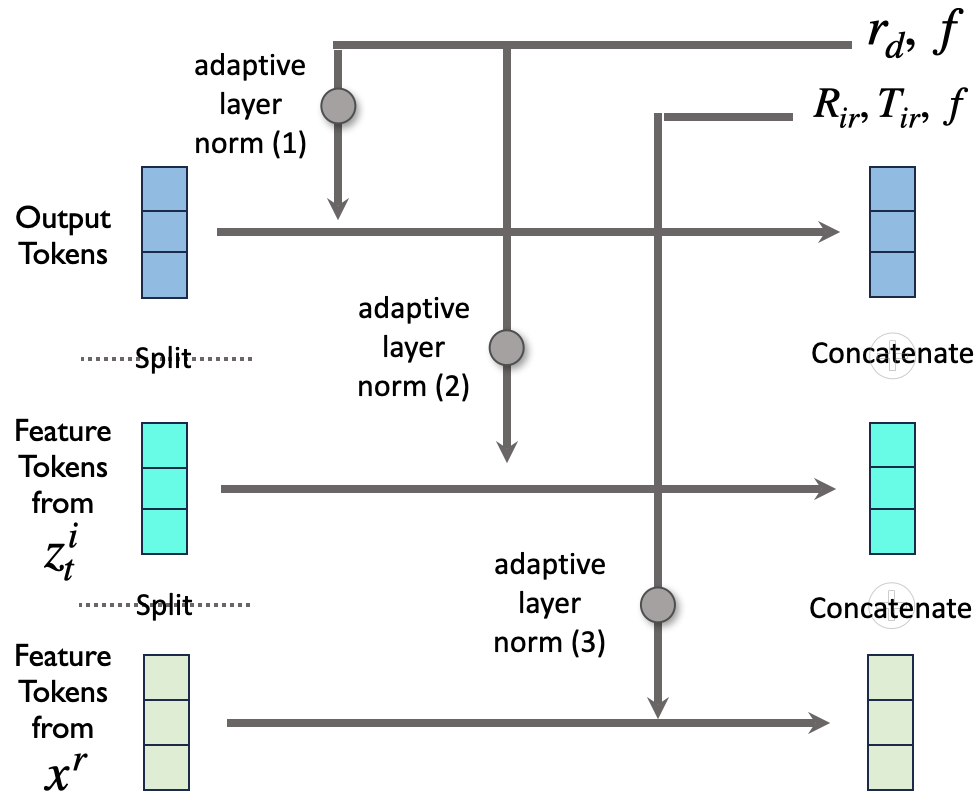}
    \caption{\textbf{Camera Conditioning: } We apply AdaLN separately for output tokens, feature tokens from noisy input image and those from reference image. As input rotation matrices are always identity, so we condition on camera distance $r_d$ and focal length $f$ on output tokens and feature tokens from $z_t^i$ with different embedding MLPs. We condition feature tokens from $x^r$ on relative pose $R_{ir}$, $T_{ir}$ between input camera pose $c^i$ and reference camera pose $c^r$.}
    \label{fig:cam_cond}
\end{wrapfigure}

\subsection{Predicting the scene: Transformer, Relative Pose, VM-Representation}


Applying the transformer architecture to diffusion models presents unique challenges compared to U-Net, particularly in translating 2D image features into 3D output tokens within each grid.
We propose four ways to help generate 3d outputs: conditioning on relative camera poses, a self-attention block for all tokens, randomly swapping the camera reference frame to be aligned with the input view or the reference view, applying dropout to conditioning images and leveraging a vector-matrix representation.

\textbf{Relative Pose: }To facilitate the job of the decoder, 
we assume that input views $c^i$ always have the identity rotation. 
Since MVImgNet acquires the camera pose from SfM~\cite{schoenberger2016mvs, schoenberger2016sfm} where different scenes have different scales and alignment, 
we apply an affine transformation to move the input camera to be at $(0,0,-r_d)$, with identity rotation, where $r_d$ is the distance between the centre of the coordinate system and the input camera. 
The reference image $x^r$ transforms accordingly so that it has relative pose $R_{ir}, T_{ir}$ between $c^i$ and $c^r$.

\textbf{Conditioning on Camera Parameters: }The decoder employs self-attention only, by concatenating feature tokens from the encoder with output tokens which are replicated for grid position, then differentiated by learnable positional embedding.
The decoder conditions on camera parameters using adaptive layer normalization (AdaLN)~\cite{xu2019understanding}.
Unlike DiT~\cite{peebles2023scalable} which conditions diffusion steps via AdaLN, we condition camera parameters only.
Within each attention block, we split the tokens as shown in Figure~\ref{fig:cam_cond}, and apply AdaLN separately following their relative camera pose.
This allows the decoder to condition on camera parameters better as shown in Table~\ref{tab:quantitative_results}.

\textbf{Swapping input views and reference views: }
The model tended to stick to a degraded solution when the noisy images were provided as input throughout the training, as shown in Table ~\ref{tab:quantitative_results}.
To address this, We randomly swap the positions of noisy input images with reference images during the training, effectively placing the reference image with identity rotation.
Additionally, we apply dropout to reference images, which regularizes the model and enables it to perform unconditional generation and classifier-free guidance.

\textbf{VM Representation: } Representing the scene as a voxel-grid is computationally expensive. 
Unlike 2D matrices, there is no Eckart-Young Theorem in tensor decomposition.
We adopt the Vector-Matrix Representation (VM Representation) proposed by TensoRF~\cite{chen2022tensorf}.
The idea is to decompose a 3D tensor $\mathcal{S}$ into the summation of three matrices $M^{Y,Z}, M^{Z,X}, M^{X,Y}$ and vectors $V^X, V^Y, V^Z$ with $k$ number of channels.

\begin{equation}
    \hat{\mathcal{S}} = \sum_{r_1=1}^k V^X_{r_1} \circ M_{r_1}^{Y,Z} + \sum_{r_2=1}^k V^Y_{r_2} \circ M_{r_2}^{Z,X} + \sum_{r_3=1}^k V_{r_3}^Z \circ M_{r_3}^{X,Y}.
    \label{eq:vm_representation}
\end{equation}

The difference between \OURS{} and TensoRF is that each grid in \OURS{} communicates via attention mechanism within the decoder, leading to a more compact and efficient representation. 
After the decoder, \OURS{} embeds and reshapes the output tensors to build the VM representation. Finally, points are queried along the ray computed from the camera poses $c^i$ and $c^n$, both are affine-transformed as $c^i$ is the identity rotation.

\subsection{Rendering}

\begin{figure}[tb]
    \centering
    \hspace*{-4mm}\includegraphics[width=1.05\textwidth]{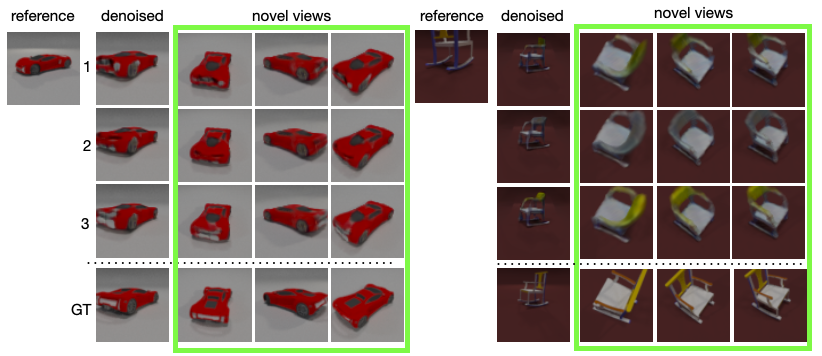}
    \caption{\textbf{Sampling results on ShapeNet: } 
    Given reference images, we show different sampling results at input viewpoint (denoised), \textit{e.g.}, identity pose, and 
    novel view synthesis results for objects of the car and chair category from ShapeNet. 
    }
    \label{fig:shapenet_diverse}
\end{figure}

Once we build the estimated scene $\hat{S}$ as a vector-matrix representation, we can render from any viewpoint.
From camera poses $c^i$ and $c^n$, we sample the points $q_j$ along the ray $r$ and query them to $\hat{\mathcal{S}}$ which is composed of vectors $V_X, V_Y, V_Z$ and matrices $M_{Y,Z}, M_{Z,X}, M_{X,Y}$. 
Query points $q_j$ are projected onto three vectors and matrices, and features are computed via bilinear sampling. 
We compute the density $\sigma_u$ at $q_u$ using Equation \ref{eq:vm_representation}, and for color $\mathbf{c}_u$, we use shallow MLP to regress the dimensions into 3. 
Then, we can compute the color of the ray $\hat{C}(r)$ by volume rendering, by first computing the transmittance $T_u=\exp(-\sum_{l=1}^{u-1}\sigma_l \delta_l)$ where $\delta_l$ is the distance between adjacent query points. 

\begin{equation}
    \hat{C}(r) = \sum_{u=1}^\text{N} T_u(1-\exp(-\sigma_u\delta_u))\mathbf{c}_u
    \label{eq:volume_rendering}
\end{equation}

\subsection{Losses: Photometric, Denoising, and Sampling}

We supervise the model with the denoising loss on ${x}^i$ as 
 shown in Equation~\ref{eq:denoising}. In addition,  to make the model truly 3d-aware, we also generate novel views by rendering the vector-matrix representation from new viewpoints, and compute a photometric loss. 

Using Equation \ref{eq:volume_rendering}, we can render novel views $\hat{x}^n$ from novel viewpoints $c^n$, as well as the input view $\hat{x}^i$ as shown below.

\begin{equation}
    \hat{x}^n, \hat{x}^i = \mathcal{F}_\theta(z_t^i, x^r, R_{ir}, T_{ir}; c^i, c^r, c^n)
\end{equation}

We employ a photometric loss by computing the $L_2$ loss that measures the discrepancy between the ground truth $x^n$ and rendered views $\hat{x}^n$. 

%
\begin{equation}
    \mathcal{L}_{photo} = \mathbb{E}_{\theta,t,x^r}[(\hat{x}^n - x^n)^2] 
\end{equation}

We additionally use the LPIPS~\cite{zhang2018unreasonable} loss on novel views $\hat{x}^n$ as well as the input view $\hat{x}^i$. To regularize the model further, we also employ the distortion loss $\mathcal{L}_{dist}$ proposed in  MipNeRF360 ~\cite{barron2022mip}.
The total loss function is as below.
\begin{equation}
    \mathcal{L}_{total} = \mathcal{L}_{denoising} + \gamma_1\mathcal{L}_{photo} + \gamma_2\mathcal{L}_{\textsc{\tiny{LPIPS}}}+ \gamma_3\mathcal{L}_{dist}
\end{equation}

\noindent\textbf{Sampling:} In 3D implicit representation, we only observe the noise level from the specific viewpoint. As we take the relative-pose-based approach, we denoise from the input viewpoint, which has identity rotation matrix. During the inference time, we use DDIM sampling~\cite{song2020denoising}, see Figure~\ref{fig:shapenet_diverse} for examples of sampling results.




\section{Experimental Evaluation}

\begin{table}
    \centering
        \caption{\textbf{Quantitative Results on MVimgNet : } \OURS{} achieves similar performance with NViST, and performs better on FID. 
        This table also present our ablation study on MVImgNet Landscape dataset. 
        Note that we compute FID for ours and $\epsilon$-parameterization, as others do not perform well on other metrics.
        We also present GIBR~\cite{anciukevicius2024denoising} result, though their resolution is 256$\times$256.
        }
        \begin{tabular}{@{}lcccc@{}}
            \toprule
             & \multicolumn{4}{c}{MVImgNet (Landscape / Portrait)} \\ \midrule
             &  \multicolumn{1}{c}{PSNR$\uparrow$} & \multicolumn{1}{c}{SSIM$\uparrow$} & \multicolumn{1}{c}{LPIPS$\downarrow$} & \multicolumn{1}{c}{FID$\downarrow$}  \\ \midrule

            Ours &   20.82 / 20.65 &  0.62 / \textbf{0.63} &   0.22 / 0.21  &   37.8 / 14.32 \\ 
            Ours (w/cfg 2.0)  & 20.91 / 20.85 & \textbf{0.63} / \textbf{0.63}  & 0.21 / \textbf{0.20}   &  \textbf{31.82} / \textbf{11.68}  \\
            NViST~\cite{jang2023nvist} & \textbf{21.03} / \textbf{21.23}  &  0.62 / 0.62  &  \textbf{0.20} / 0.21  & 39.35 / 16.68   \\ \midrule
            w/ Cross-Attn & \multicolumn{4}{c}{did not converge} \\
            w/ noisy input only &  18.62 & 0.41  & 0.30  &  --\\
            w/o Encoder &  17.32 &  0.42  & 0.35 &   --\\
            w/o Decoder Camera Conditioning &  17.19 & 0.38 & 0.33  & -- \\
            w/ $\epsilon$-parameterization (w/cfg 2.0) & 20.72 &  0.61  & 0.22 & 44.2 \\
            \midrule
            GIBR~\cite{anciukevicius2024denoising} (higher resolution) &  17.96  & 0.554 &  0.519 & 107.3 \\ \bottomrule
        \end{tabular}
    \label{tab:quantitative_results}
\end{table}

\subsection{MVImgNet}

\begin{figure}[tb]
    \centering
    \includegraphics[width=\textwidth]{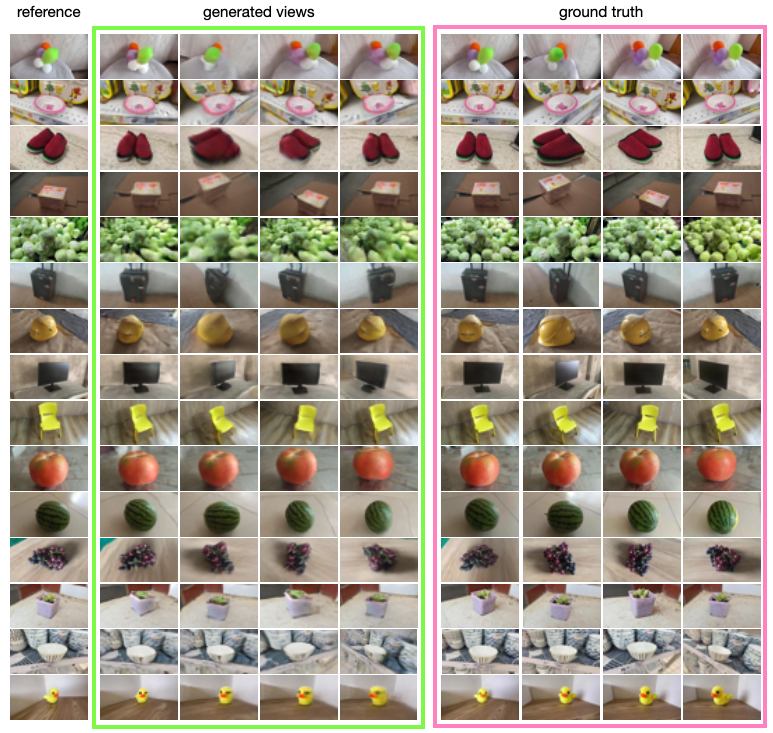}
    \caption{\textbf{Qualitative Results on MVImgNet: } We show the capabilities of \OURS{} on test images from unknown scenes of MVImgNet~\cite{yu2023mvimgnet}. The model takes a reference image as input and denoises at input viewpoint during sampling.
    We show novel view synthesis results for three different viewpoints after the denoising step.}
    \label{fig:qualitative_landscape}
\end{figure}

\begin{figure}[tb]
    \includegraphics[width=\textwidth]{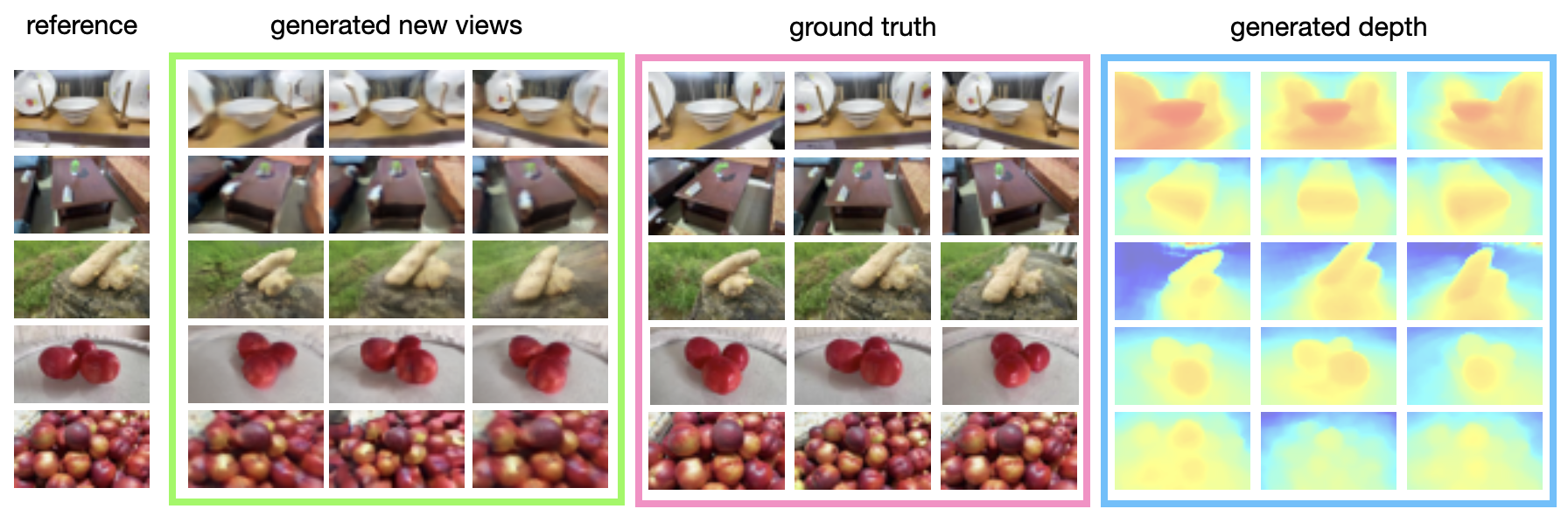}
    \caption{\textbf{Qualitative Results with depth prediction: } We show reference images, generated new views and the  generated depth maps for a variety of scenes from MVImgNet, as well as a comparison with the ground truth views. 
    } 
    \label{fig:qualitative_landscape_depth}
    \vspace{0.0cm}
\end{figure}

\noindent\textbf{Dataset: } MVImgNet consists of 6.5M images from 220K scenes across 238 cateogories, all of them are real world captures, and the camera poses are computed through COLMAP~\cite{schoenberger2016sfm, schoenberger2016mvs}, and each scene usually contains around 30 images.
The dataset contains both portrait and aspect ratio, and we train the model separately for each aspect ratio. 
We split train/test for holding out every 50 scenes in alphabetical order.
We downsample and center-crop images to 56$\times$32 and 32$\times$56, 
we also downscale the point clouds from COLMAP to unit-cube, and change focal length accordingly.
See Figure~\ref{fig:qualitative_landscape} for qualitative results and Figure~\ref{fig:qualitative_landscape_depth} for depth predictions.

\noindent\textbf{Baselines: } We train NViST~\cite{jang2023nvist}, which is a deterministic approach for training multiple unaligned real-world captures using the MVImgNet dataset.
For the diffusion-based model, many approaches assume that scenes are aligned~\cite{szymanowicz2023viewset, karnewar2023holodiffusion, anciukevivcius2023renderdiffusion}, but they would not be appropriate for this dataset which is not aligned. 
We quantitatively compare with GIBR~\cite{anciukevicius2024denoising} on MVImgNet as in Table~\ref{tab:quantitative_results}, but note that they use a different train/test split, and a different resolutions.

\begin{figure}[tb]
    \includegraphics[width=\textwidth]{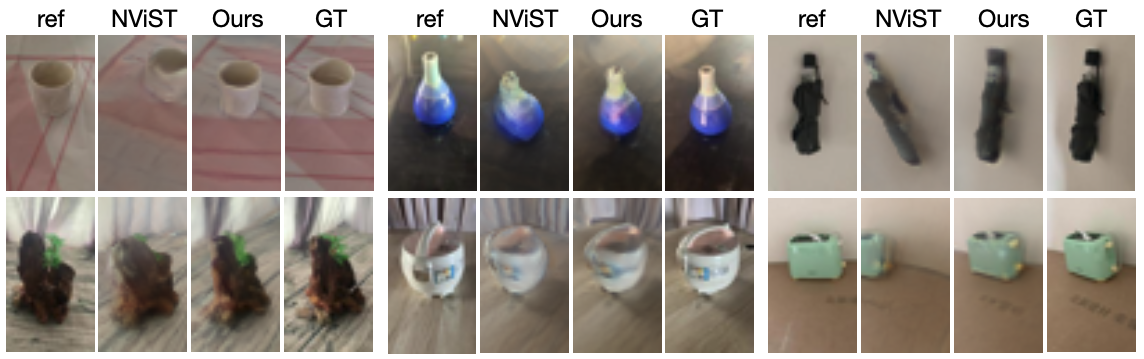}
    \caption{\textbf{Qualitative Comparison: }We compare \OURS{} with NViST~\cite{jang2023nvist}. For NViST, we provide the same reference images as input images. NViST sometimes fails to understand the scale and pose, and also generates more blurry images.} 
    \label{fig:qualitative_comparison}
\end{figure}


\noindent\textbf{Results: } Table~\ref{tab:quantitative_results} presents our results.
\OURS{} achieves better performance than NViST on FID, 
and performs similar to NViST on LPIPS~\cite{zhang2018unreasonable} and SSIM~\cite{wang2004image}. 
Note that NViST minimizes the L2 distance, and it may contribute to the better results on these metrics.
Qualitatively, \OURS{} generates sharper images, and NViST sometimes struggles to estimate the scale of scenes with compared to \OURS{}, see Figure~\ref{fig:qualitative_comparison}.
Moreover our method outperforms GIBR on FID~\cite{Heusel2017GANsTB}, 
although, it is worth noting that their method uses a different train/test split, and uses the different resolutions. 
GIBR cropped and downsampled images to 90$\times$90 or 256$\times$256, while \OURS{} pre-processes to 32$\times$56 or 56$\times$32.

\subsection{Ablation Study}

In order to develop \OURS{} we made different design choices, such as not using cross-attention. 
The lower part of Table~\ref{tab:quantitative_results} presents different design variations and we discuss these variations as follow: 

\textbf{Cross-Attention based model: } NViST~\cite{jang2023nvist} and LRM~\cite{hong2023lrm}, which are both deterministic approaches, employ cross-attention and self-attention architectures in their decoders.
The primary difference between self-attention and cross-attention is that feature tokens (output from the encoder) from noisy or reference images are not updated in the decoder, leading to smaller attention matrices. However, our training did not converge when we applied cross and self-attention approach to the decoder.

\textbf{Fixing noisy images at input view: } During the inference, we denoise from the input view (\textit{i.e.,} identity rotation). 
If we always feed the noisy images with identity rotation during training, the model tends to find a trivial solution instead of learning the 3D structure from multiple views, which leads to lower performance. 
Randomly swapping the reference frame between the reference image and the sampled noisy input image regularizes the model for better performance, as shown in Table~\ref{tab:quantitative_results}.

\textbf{$\epsilon$-parameterization:} DDPM~\cite{ho2020denoising} proposed using $\epsilon$-parameterization, and Hang \textit{et al.}~\cite{hang2023efficient} suggested applying the minimum Signal Noise Ratio (SNR) to $x$-parameterization, resulting in a model that converges similarly to $\epsilon$-parameterization. 
Here, we use the weight min(SNR(t) + 1, 5) for $x$-parameterization, which is akin to $v$-parameterization. In our ablation study, we compare models using the weights min(SNR(t), 5) and min(SNR(t) + 1, 5), corresponding to $\epsilon$-parameterization and $v$-parameterization, respectively. We found that with the weight min(SNR(t), 5), the model performance degrades in terms of FID score.

\textbf{Without Encoder:} Since we employ self-attention in our Transformers, we conducted an ablation study to determine the role of the encoder. 
Without the encoder, the model's performance significantly drops. 

\textbf{No Camera Conditioning on Decoder:} To test the effectiveness of our camera conditioning approach depicted in Figure~\ref{fig:cam_cond}, we replaced our decoder with ViT. 
In this setup, the decoder needs to infer the camera pose from feature image tokens alone. 
As shown in Table~\ref{tab:quantitative_results}, the model fails to train properly without conditioning on camera parameters.

\begin{table}
    \centering
    \caption{
        \textbf{Quantitative Results on ShapeNet: } \OURS{} performs similar or better than baseline models on all ShapeNet categories. Our model performs much better than other diffusion-based models (RD, GIBR, VSD) in terms of FID score, and achieves similar FID score compared to EG3D which is 3D-aware GAN~\cite{eg3dChan2021}.
        We denote split from RD and GIBR with $^1$ and $^2$. 
    }
    \label{tab:shapenet} 
    \begin{tabular}{@{}lccc@{}}
        \toprule
         & \multicolumn{1}{c}{Cars}  &\multicolumn{1}{c}{Planes} &\multicolumn{1}{c}{Chairs} \\ \midrule
         &\small{PSNR$\uparrow$/SSIM$\uparrow$/LPIPS$\downarrow$/FID$\downarrow$} 
         &\small{PSNR$\uparrow$/SSIM$\uparrow$/LPIPS$\downarrow$/FID$\downarrow$} 
         &\small{PSNR$\uparrow$/SSIM$\uparrow$/LPIPS$\downarrow$/FID$\downarrow$} 
         \\ 
        \midrule
    Ours &  27.64 / 0.87 / \textbf{0.12}/ \textbf{14.9} & \textbf{26.9} / \textbf{0.88} / 0.15 / 21.8 &  \textbf{28.75} / \textbf{0.88} / 0.12 / 15.6 \\ 
    \midrule
    RD$^{1}$~\cite{anciukevivcius2023renderdiffusion} & 25.4 / 0.81 / -- / 46.5  & 26.3 / 0.83 / -- / 53.3  & 26.6 / 0.83 / -- / 47.8 \\ 
    GIBR$^{2}$~\cite{anciukevicius2024denoising} & \textbf{29.74} / \textbf{0.90} /0.14 / 90.1 & (N/A) & (N/A) \\ 
    VSD$^{2}$~\cite{szymanowicz2023viewset} & 28.00 / 0.87 / 0.17 / 56.0 & (N/A) & (N/A) \\ 
    \midrule
    EG3D$^{1}$~\cite{eg3dChan2021} & 21.8 / 0.71 / -- / 17.9 & 25.0 / 0.80 / -- / \textbf{20.9} & 25.5 / 0.80 / -- / \textbf{14.2} \\
        \bottomrule
    \end{tabular}
\end{table}

\subsection{ShapeNet}

\textbf{Dataset:} We use ShapeNet renderings from~\cite{anciukevivcius2023renderdiffusion} to validate our model. 
The dataset includes three categories (cars, chairs and planes) with each category containing 3,200 scenes, divided into training (2700) and testing (500) sets. 
The dataset assumes an object-centric scene, meaning the objects are always located at the center of the coordinate system. 
It also adopts the simplified camera model which always points toward the center of the coordinate system. 
Additionally, all 3D objects are aligned, sharing the same reference frame, \textit{e.g.}, plane nose aligned with the positive x-axis. 
This setting differs significantly from MVImgNet~\cite{yu2023mvimgnet}, where all scenes are not aligned and assume a 6-degree-of-freedom camera model.

\textbf{Baselines and Results:} Even with aligned scenes, we adopt the relative-pose-based approach, meaning we do not exploit the 3D alignment of objects in this dataset. 
We follow the training protocol from RenderDiffusion and compare our results with RenderDiffusion, GIBR, Viewset Diffusion~\cite{szymanowicz2023viewset}, and the 3D-aware GAN EG3D~\cite{eg3dChan2021}. 
As shown in Table~\ref{tab:shapenet}, \OURS{} performs similarly or better across most metrics. 
Notably, \OURS{} excels in terms of the FID score compared to other diffusion models, achieving a level better or comparable to EG3D.


\textbf{Diverse Sampling:} We validate our model's capability of generating multiple outputs for occluded regions from a single image, as shown in Figure~\ref{fig:shapenet_diverse}. 
We deliberately choose viewpoints that are not visible in the reference image and denoise from there. The model demonstrates that it can generate multiple plausible outputs, maintaining consistency across viewpoints.

\subsection{Implementation Details}


We train our model using 2 A100-40GB GPUs for both MVImgNet and ShapeNet datasets, using the same architecture with both.
Training MVImgNet takes 5 days (700,000 iterations) for both landscape and portrait, while training ShapeNet takes 2 days (400,000 iterations). 
We use a batch size of 44 for MVImgNet and 26 for ShapeNet. 
We use AdamW~\cite{loshchilov2019decoupled} optimizer and the learning rate is set to 2e-4 with cosine decay, and a warm-up period of 50,000 iterations is applied for both datasets. 
\section{Conclusion}

We have introduced \OURS{}, a transformer-based 3D-aware diffusion model trained on real-world multiview images. Our evaluation demonstrates that \OURS{} outperforms baseline models, particularly in terms of FID score. We show qualitative results on a challenging real-world dataset of casually captured videos of everyday scenes (MVImgNet) that demonstrate the abitity of \OURS{} to generate novel views of complex scenes of a large variety of object categories with very different backgrounds. A detailed ablation study is provided to illustrate the successful training of this model. An interesting future direction includes extending this model to more challenging outdoor scenes, higher resolution or associating with other modalities.

\textbf{Limitation:} We needed to downsample images significantly to train this model, resulting in a loss of output quality. This was because we only had access to two A100 GPUs. The model also occasionally struggles with outdoor scenes, partly due to the VM representation. Recent diffusion approaches, like flow models~\cite{lipman2022flow}, could not be employed. This is because they predict velocity (defined as "noise - ground truth"), but we cannot observe this noise for novel view synthesis. Instead, we apply a mathematical equivalent of $v$-parameterization when we supervise the model with an L2 loss and other auxiliary losses after volume rendering.


\bibliographystyle{plain}
\bibliography{references}

\begin{thebibliography}{10}

\bibitem{anciukevicius2024denoising}
Titas Anciukevicius, Fabian Manhardt, Federico Tombari, and Paul Henderson.
\newblock Denoising diffusion via image-based rendering.
\newblock {\em arXiv preprint arXiv:2402.03445}, 2024.

\bibitem{anciukevivcius2023renderdiffusion}
Titas Anciukevi{\v{c}}ius, Zexiang Xu, Matthew Fisher, Paul Henderson, Hakan Bilen, Niloy~J Mitra, and Paul Guerrero.
\newblock Renderdiffusion: Image diffusion for 3d reconstruction, inpainting and generation.
\newblock In {\em Proceedings of the IEEE/CVF Conference on Computer Vision and Pattern Recognition}, pages 12608--12618, 2023.

\bibitem{athar2022rignerf}
ShahRukh Athar, Zexiang Xu, Kalyan Sunkavalli, Eli Shechtman, and Zhixin Shu.
\newblock Rignerf: Fully controllable neural 3d portraits.
\newblock In {\em Computer Vision and Pattern Recognition (CVPR)}, 2022.

\bibitem{barron2022mip}
Jonathan~T Barron, Ben Mildenhall, Dor Verbin, Pratul~P Srinivasan, and Peter Hedman.
\newblock Mip-nerf 360: Unbounded anti-aliased neural radiance fields.
\newblock In {\em Proceedings of the IEEE/CVF Conference on Computer Vision and Pattern Recognition}, pages 5470--5479, 2022.

\bibitem{cai2022pix2nerf}
Shengqu Cai, Anton Obukhov, Dengxin Dai, and Luc Van~Gool.
\newblock Pix2nerf: Unsupervised conditional p-gan for single image to neural radiance fields translation.
\newblock In {\em Proceedings of the IEEE/CVF conference on computer vision and pattern recognition}, pages 3981--3990, 2022.

\bibitem{chanmonteiro2020pi-GAN}
Eric Chan, Marco Monteiro, Petr Kellnhofer, Jiajun Wu, and Gordon Wetzstein.
\newblock pi-gan: Periodic implicit generative adversarial networks for 3d-aware image synthesis.
\newblock In {\em arXiv}, 2020.

\bibitem{eg3dChan2021}
Eric~R. Chan, Connor~Z. Lin, Matthew~A. Chan, Koki Nagano, Boxiao Pan, Shalini~De Mello, Orazio Gallo, Leonidas Guibas, Jonathan Tremblay, Sameh Khamis, Tero Karras, and Gordon Wetzstein.
\newblock Efficient geometry-aware {3D} generative adversarial networks.
\newblock In {\em arXiv}, 2021.

\bibitem{chan2023genvs}
Eric~R Chan, Koki Nagano, Matthew~A Chan, Alexander~W Bergman, Jeong~Joon Park, Axel Levy, Miika Aittala, Shalini De~Mello, Tero Karras, and Gordon Wetzstein.
\newblock Genvs: Generative novel view synthesis with 3d-aware diffusion models.
\newblock {\em arXiv preprint arXiv:2311.10709}, 2023.

\bibitem{chen2022tensorf}
Anpei Chen, Zexiang Xu, Andreas Geiger, Jingyi Yu, and Hao Su.
\newblock Tensorf: Tensorial radiance fields.
\newblock In {\em European Conference on Computer Vision}, pages 333--350. Springer, 2022.

\bibitem{chen2021mvsnerf}
Anpei Chen, Zexiang Xu, Fuqiang Zhao, Xiaoshuai Zhang, Fanbo Xiang, Jingyi Yu, and Hao Su.
\newblock Mvsnerf: Fast generalizable radiance field reconstruction from multi-view stereo.
\newblock In {\em Proceedings of the IEEE/CVF international conference on computer vision}, pages 14124--14133, 2021.

\bibitem{chen2023matchnerf}
Yuedong Chen, Haofei Xu, Qianyi Wu, Chuanxia Zheng, Tat-Jen Cham, and Jianfei Cai.
\newblock Explicit correspondence matching for generalizable neural radiance fields.
\newblock {\em arXiv preprint arXiv:2304.12294}, 2023.

\bibitem{dhariwal2021diffusion}
Prafulla Dhariwal and Alexander Nichol.
\newblock Diffusion models beat gans on image synthesis.
\newblock {\em Advances in neural information processing systems}, 34:8780--8794, 2021.

\bibitem{dosovitskiy2020image}
Alexey Dosovitskiy, Lucas Beyer, Alexander Kolesnikov, Dirk Weissenborn, Xiaohua Zhai, Thomas Unterthiner, Mostafa Dehghani, Matthias Minderer, Georg Heigold, Sylvain Gelly, et~al.
\newblock An image is worth 16x16 words: Transformers for image recognition at scale.
\newblock {\em arXiv preprint arXiv:2010.11929}, 2020.

\bibitem{fridovich2023k}
Sara Fridovich-Keil, Giacomo Meanti, Frederik~Rahb{\ae}k Warburg, Benjamin Recht, and Angjoo Kanazawa.
\newblock K-planes: Explicit radiance fields in space, time, and appearance.
\newblock In {\em Proceedings of the IEEE/CVF Conference on Computer Vision and Pattern Recognition}, pages 12479--12488, 2023.

\bibitem{Gafni_2021_CVPR}
Guy Gafni, Justus Thies, Michael Zollh{\"o}fer, and Matthias Nie{\ss}ner.
\newblock Dynamic neural radiance fields for monocular 4d facial avatar reconstruction.
\newblock In {\em Proceedings of the IEEE/CVF Conference on Computer Vision and Pattern Recognition (CVPR)}, pages 8649--8658, June 2021.

\bibitem{gu2023nerfdiff}
Jiatao Gu, Alex Trevithick, Kai-En Lin, Joshua~M Susskind, Christian Theobalt, Lingjie Liu, and Ravi Ramamoorthi.
\newblock Nerfdiff: Single-image view synthesis with nerf-guided distillation from 3d-aware diffusion.
\newblock In {\em Proceedings of the International Conference on Machine Learning}, 2023.

\bibitem{hang2023efficient}
Tiankai Hang, Shuyang Gu, Chen Li, Jianmin Bao, Dong Chen, Han Hu, Xin Geng, and Baining Guo.
\newblock Efficient diffusion training via min-snr weighting strategy.
\newblock In {\em Proceedings of the IEEE/CVF International Conference on Computer Vision}, pages 7441--7451, 2023.

\bibitem{he2022masked}
Kaiming He, Xinlei Chen, Saining Xie, Yanghao Li, Piotr Doll{\'a}r, and Ross Girshick.
\newblock Masked autoencoders are scalable vision learners.
\newblock In {\em Proceedings of the IEEE/CVF conference on computer vision and pattern recognition}, pages 16000--16009, 2022.

\bibitem{Henzler_2021_CVPR}
Philipp Henzler, Jeremy Reizenstein, Patrick Labatut, Roman Shapovalov, Tobias Ritschel, Andrea Vedaldi, and David Novotny.
\newblock Unsupervised learning of 3d object categories from videos in the wild.
\newblock In {\em Proceedings of the IEEE/CVF Conference on Computer Vision and Pattern Recognition (CVPR)}, pages 4700--4709, June 2021.

\bibitem{Heusel2017GANsTB}
Martin Heusel, Hubert Ramsauer, Thomas Unterthiner, Bernhard Nessler, and Sepp Hochreiter.
\newblock Gans trained by a two time-scale update rule converge to a local nash equilibrium.
\newblock {\em arXiv preprint arXiv:1706.08500}, 2017.

\bibitem{ho2022imagen}
Jonathan Ho, William Chan, Chitwan Saharia, Jay Whang, Ruiqi Gao, Alexey Gritsenko, Diederik~P Kingma, Ben Poole, Mohammad Norouzi, David~J Fleet, et~al.
\newblock Imagen video: High definition video generation with diffusion models.
\newblock {\em arXiv preprint arXiv:2210.02303}, 2022.

\bibitem{ho2020denoising}
Jonathan Ho, Ajay Jain, and Pieter Abbeel.
\newblock Denoising diffusion probabilistic models.
\newblock {\em Advances in neural information processing systems}, 33:6840--6851, 2020.

\bibitem{ho2022classifier}
Jonathan Ho and Tim Salimans.
\newblock Classifier-free diffusion guidance.
\newblock {\em arXiv preprint arXiv:2207.12598}, 2022.

\bibitem{hong2022headnerf}
Yang Hong, Bo~Peng, Haiyao Xiao, Ligang Liu, and Juyong Zhang.
\newblock Headnerf: A real-time nerf-based parametric head model.
\newblock In {\em Proceedings of the IEEE/CVF Conference on Computer Vision and Pattern Recognition}, pages 20374--20384, 2022.

\bibitem{hong2023lrm}
Yicong Hong, Kai Zhang, Jiuxiang Gu, Sai Bi, Yang Zhou, Difan Liu, Feng Liu, Kalyan Sunkavalli, Trung Bui, and Hao Tan.
\newblock Lrm: Large reconstruction model for single image to 3d.
\newblock {\em arXiv preprint arXiv:2311.04400}, 2023.

\bibitem{irshad2023neo360}
Muhammad~Zubair Irshad, Sergey Zakharov, Katherine Liu, Vitor Guizilini, Thomas Kollar, Adrien Gaidon, Zsolt Kira, and Rares Ambrus.
\newblock Neo 360: Neural fields for sparse view synthesis of outdoor scenes.
\newblock 2023.

\bibitem{jang2021codenerf}
Wonbong Jang and Lourdes Agapito.
\newblock Codenerf: Disentangled neural radiance fields for object categories.
\newblock In {\em Proceedings of the IEEE/CVF International Conference on Computer Vision}, pages 12949--12958, 2021.

\bibitem{jang2023nvist}
Wonbong Jang and Lourdes Agapito.
\newblock Nvist: In the wild new view synthesis from a single image with transformers.
\newblock {\em arXiv preprint arXiv:2312.08568}, 2023.

\bibitem{jun2023shap}
Heewoo Jun and Alex Nichol.
\newblock Shap-e: Generating conditional 3d implicit functions.
\newblock {\em arXiv preprint arXiv:2305.02463}, 2023.

\bibitem{karnewar2023holodiffusion}
Animesh Karnewar, Andrea Vedaldi, David Novotny, and Niloy~J Mitra.
\newblock Holodiffusion: Training a 3d diffusion model using 2d images.
\newblock In {\em Proceedings of the IEEE/CVF Conference on Computer Vision and Pattern Recognition}, pages 18423--18433, 2023.

\bibitem{kerbl3Dgaussians}
Bernhard Kerbl, Georgios Kopanas, Thomas Leimk{\"u}hler, and George Drettakis.
\newblock 3d gaussian splatting for real-time radiance field rendering.
\newblock {\em ACM Transactions on Graphics}, 42(4), July 2023.

\bibitem{kim2023nfldm}
Seung~Wook Kim, Bradley Brown, Kangxue Yin, Karsten Kreis, Katja Schwarz, Daiqing Li, Robin Rombach, Antonio Torralba, and Sanja Fidler.
\newblock Neuralfield-ldm: Scene generation with hierarchical latent diffusion models.
\newblock In {\em IEEE Conference on Computer Vision and Pattern Recognition ({CVPR})}, 2023.

\bibitem{kulhanek2022viewformer}
Jon{\'a}{\v{s}} Kulh{\'a}nek, Erik Derner, Torsten Sattler, and Robert Babu{\v{s}}ka.
\newblock Viewformer: Nerf-free neural rendering from few images using transformers.
\newblock In {\em Computer Vision--ECCV 2022: 17th European Conference, Tel Aviv, Israel, October 23--27, 2022, Proceedings, Part XV}, pages 198--216. Springer, 2022.

\bibitem{le2022stylemorph}
Eric-Tuan Le, Edward Bartrum, and Iasonas Kokkinos.
\newblock Stylemorph: Disentangled 3d-aware image synthesis with a 3d morphable stylegan.
\newblock In {\em The Eleventh International Conference on Learning Representations}, 2022.

\bibitem{lipman2022flow}
Yaron Lipman, Ricky~TQ Chen, Heli Ben-Hamu, Maximilian Nickel, and Matt Le.
\newblock Flow matching for generative modeling.
\newblock {\em arXiv preprint arXiv:2210.02747}, 2022.

\bibitem{liu2023zero}
Ruoshi Liu, Rundi Wu, Basile Van~Hoorick, Pavel Tokmakov, Sergey Zakharov, and Carl Vondrick.
\newblock Zero-1-to-3: Zero-shot one image to 3d object.
\newblock In {\em Proceedings of the IEEE/CVF International Conference on Computer Vision}, pages 9298--9309, 2023.

\bibitem{loshchilov2019decoupled}
Ilya Loshchilov and Frank Hutter.
\newblock Decoupled weight decay regularization.
\newblock In {\em International Conference on Learning Representations}, 2019.

\bibitem{melas2023realfusion}
Luke Melas-Kyriazi, Iro Laina, Christian Rupprecht, and Andrea Vedaldi.
\newblock Realfusion: 360deg reconstruction of any object from a single image.
\newblock In {\em Proceedings of the IEEE/CVF Conference on Computer Vision and Pattern Recognition}, 2023.

\bibitem{mildenhall2021nerf}
Ben Mildenhall, Pratul~P Srinivasan, Matthew Tancik, Jonathan~T Barron, Ravi Ramamoorthi, and Ren Ng.
\newblock Nerf: Representing scenes as neural radiance fields for view synthesis.
\newblock {\em Communications of the ACM}, 65(1):99--106, 2021.

\bibitem{Miyato2024GTA}
Takeru Miyato, Bernhard Jaeger, Max Welling, and Andreas Geiger.
\newblock Gta: A geometry-aware attention mechanism for multi-view transformers.
\newblock In {\em International Conference on Learning Representations (ICLR)}, 2024.

\bibitem{muller2022autorf}
Norman M{\"u}ller, Andrea Simonelli, Lorenzo Porzi, Samuel~Rota Bul{\`o}, Matthias Nie{\ss}ner, and Peter Kontschieder.
\newblock Autorf: Learning 3d object radiance fields from single view observations.
\newblock In {\em Proceedings of the IEEE/CVF Conference on Computer Vision and Pattern Recognition}, pages 3971--3980, 2022.

\bibitem{muller2022instant}
Thomas M{\"u}ller, Alex Evans, Christoph Schied, and Alexander Keller.
\newblock Instant neural graphics primitives with a multiresolution hash encoding.
\newblock {\em ACM Transactions on Graphics (ToG)}, 41(4):1--15, 2022.

\bibitem{niemeyer2020differentiable}
Michael Niemeyer, Lars Mescheder, Michael Oechsle, and Andreas Geiger.
\newblock Differentiable volumetric rendering: Learning implicit 3d representations without 3d supervision.
\newblock In {\em Proceedings of the IEEE/CVF Conference on Computer Vision and Pattern Recognition (CVPR)}, pages 3504--3515. IEEE, 2020.

\bibitem{peebles2023scalable}
William Peebles and Saining Xie.
\newblock Scalable diffusion models with transformers.
\newblock In {\em Proceedings of the IEEE/CVF International Conference on Computer Vision}, pages 4195--4205, 2023.

\bibitem{poole2022dreamfusion}
Ben Poole, Ajay Jain, Jonathan~T Barron, and Ben Mildenhall.
\newblock Dreamfusion: Text-to-3d using 2d diffusion.
\newblock In {\em International Conference on Learning Representations}, 2022.

\bibitem{rebain2022lolnerf}
Daniel Rebain, Mark Matthews, Kwang~Moo Yi, Dmitry Lagun, and Andrea Tagliasacchi.
\newblock Lolnerf: Learn from one look.
\newblock In {\em Proceedings of the IEEE/CVF Conference on Computer Vision and Pattern Recognition}, pages 1558--1567, 2022.

\bibitem{reizenstein2021common}
Jeremy Reizenstein, Roman Shapovalov, Philipp Henzler, Luca Sbordone, Patrick Labatut, and David Novotny.
\newblock Common objects in 3d: Large-scale learning and evaluation of real-life 3d category reconstruction.
\newblock In {\em Proceedings of the IEEE/CVF International Conference on Computer Vision}, pages 10901--10911, 2021.

\bibitem{rombach2022high}
Robin Rombach, Andreas Blattmann, Dominik Lorenz, Patrick Esser, and Bj{\"o}rn Ommer.
\newblock High-resolution image synthesis with latent diffusion models.
\newblock In {\em Proceedings of the IEEE/CVF Conference on Computer Vision and Pattern Recognition (CVPR)}, pages 10684--10695. IEEE, 2022.

\bibitem{sajjadi2022scene}
Mehdi~SM Sajjadi, Henning Meyer, Etienne Pot, Urs Bergmann, Klaus Greff, Noha Radwan, Suhani Vora, Mario Lu{\v{c}}i{\'c}, Daniel Duckworth, Alexey Dosovitskiy, et~al.
\newblock Scene representation transformer: Geometry-free novel view synthesis through set-latent scene representations.
\newblock In {\em Proceedings of the IEEE/CVF Conference on Computer Vision and Pattern Recognition}, pages 6229--6238, 2022.

\bibitem{salimans2022progressive}
Tim Salimans and Jonathan Ho.
\newblock Progressive distillation for fast sampling of diffusion models.
\newblock {\em arXiv preprint arXiv:2202.00512}, 2022.

\bibitem{sargent2023vq3d}
Kyle Sargent, Jing~Yu Koh, Han Zhang, Huiwen Chang, Charles Herrmann, Pratul Srinivasan, Jiajun Wu, and Deqing Sun.
\newblock Vq3d: Learning a 3d-aware generative model on imagenet.
\newblock {\em arXiv preprint arXiv:2302.06833}, 2023.

\bibitem{zeronvs}
Kyle Sargent, Zizhang Li, Tanmay Shah, Charles Herrmann, Hong-Xing Yu, Yunzhi Zhang, Eric~Ryan Chan, Dmitry Lagun, Li~Fei-Fei, Deqing Sun, and Jiajun Wu.
\newblock {ZeroNVS}: Zero-shot 360-degree view synthesis from a single real image.
\newblock {\em arXiv preprint arXiv:2310.17994}, 2023.

\bibitem{schoenberger2016sfm}
Johannes~Lutz Sch\"{o}nberger and Jan-Michael Frahm.
\newblock Structure-from-motion revisited.
\newblock In {\em Conference on Computer Vision and Pattern Recognition (CVPR)}, 2016.

\bibitem{schoenberger2016mvs}
Johannes~Lutz Sch\"{o}nberger, Enliang Zheng, Marc Pollefeys, and Jan-Michael Frahm.
\newblock Pixelwise view selection for unstructured multi-view stereo.
\newblock In {\em European Conference on Computer Vision (ECCV)}, 2016.

\bibitem{schwarz2020graf}
Katja Schwarz, Yiyi Liao, Michael Niemeyer, and Andreas Geiger.
\newblock Graf: Generative radiance fields for 3d-aware image synthesis.
\newblock {\em Advances in Neural Information Processing Systems}, 33:20154--20166, 2020.

\bibitem{Schwarz2024ICLR}
Katja Schwarz, Seung Wook~Kim, Jun Gao, Sanja Fidler, Andreas Geiger, and Karsten Kreis.
\newblock Wildfusion: Learning 3d-aware latent diffusion models in view space.
\newblock In {\em International Conference on Learning Representations (ICLR)}, 2024.

\bibitem{shen2023gina}
Bokui Shen, Xinchen Yan, Charles~R Qi, Mahyar Najibi, Boyang Deng, Leonidas Guibas, Yin Zhou, and Dragomir Anguelov.
\newblock Gina-3d: Learning to generate implicit neural assets in the wild.
\newblock In {\em Proceedings of the IEEE/CVF Conference on Computer Vision and Pattern Recognition}, pages 4913--4926, 2023.

\bibitem{shi2023mvdream}
Yichun Shi, Peng Wang, Jianglong Ye, Mai Long, Kejie Li, and Xiao Yang.
\newblock Mvdream: Multi-view diffusion for 3d generation.
\newblock {\em arXiv preprint arXiv:2312.02201}, 2023.

\bibitem{sitzmann2019scene}
Vincent Sitzmann, Michael Zollh{\"o}fer, and Gordon Wetzstein.
\newblock Scene representation networks: Continuous 3d-structure-aware neural scene representations.
\newblock {\em Advances in Neural Information Processing Systems}, 32, 2019.

\bibitem{sohl-dickstein2015deep}
Jascha Sohl-Dickstein, Eric~A Weiss, Niru Maheswaranathan, and Surya Ganguli.
\newblock Deep unsupervised learning using nonequilibrium thermodynamics.
\newblock In {\em Proceedings of the 32nd International Conference on Machine Learning}, pages 2256--2265. PMLR, 2015.

\bibitem{song2020denoising}
Jiaming Song, Chenlin Meng, and Stefano Ermon.
\newblock Denoising diffusion implicit models.
\newblock {\em arXiv preprint arXiv:2010.02502}, 2020.

\bibitem{szymanowicz2023viewset}
Stanislaw Szymanowicz, Christian Rupprecht, and Andrea Vedaldi.
\newblock Viewset diffusion: (0-)image-conditioned 3d generative models from 2d data.
\newblock {\em Proceedings of the IEEE/CVF International Conference on Computer Vision}, 2023.

\bibitem{tang2023dreamgaussian}
Jiaxiang Tang, Jiawei Ren, Hang Zhou, Ziwei Liu, and Gang Zeng.
\newblock Dreamgaussian: Generative gaussian splatting for efficient 3d content creation.
\newblock {\em arXiv preprint arXiv:2309.16653}, 2023.

\bibitem{tewari2023forwarddiffusion}
Ayush Tewari, Tianwei Yin, George Cazenavette, Semon Rezchikov, Joshua~B. Tenenbaum, Frédo Durand, William~T. Freeman, and Vincent Sitzmann.
\newblock Diffusion with forward models: Solving stochastic inverse problems without direct supervision.
\newblock In {\em arXiv}, 2023.

\bibitem{trevithick2021grf}
Alex Trevithick and Bo~Yang.
\newblock Grf: Learning a general radiance field for 3d representation and rendering.
\newblock In {\em Proceedings of the IEEE/CVF International Conference on Computer Vision}, pages 15182--15192, 2021.

\bibitem{wang2023morpheus}
Hengyi Wang, Jingwen Wang, and Lourdes Agapito.
\newblock Morpheus: Neural dynamic 360 $\{$$\backslash$deg$\}$ surface reconstruction from monocular rgb-d video.
\newblock {\em arXiv preprint arXiv:2312.00778}, 2023.

\bibitem{wang2021ibrnet}
Qianqian Wang, Zhicheng Wang, Kyle Genova, Pratul~P Srinivasan, Howard Zhou, Jonathan~T Barron, Ricardo Martin-Brualla, Noah Snavely, and Thomas Funkhouser.
\newblock Ibrnet: Learning multi-view image-based rendering.
\newblock In {\em Proceedings of the IEEE/CVF Conference on Computer Vision and Pattern Recognition}, pages 4690--4699, 2021.

\bibitem{wang2023prolificdreamer}
Zhengyi Wang, Cheng Lu, Yikai Wang, Fan Bao, Chongxuan Li, Hang Su, and Jun Zhu.
\newblock Prolificdreamer: High-fidelity and diverse text-to-3d generation with variational score distillation.
\newblock {\em Advances in Neural Information Processing Systems}, 2023.

\bibitem{wang2004image}
Zhou Wang, Alan~C Bovik, Hamid~R Sheikh, and Eero~P Simoncelli.
\newblock Image quality assessment: from error visibility to structural similarity.
\newblock {\em IEEE transactions on image processing}, 13(4):600--612, 2004.

\bibitem{watson2022novel}
Daniel Watson, William Chan, Ricardo Martin-Brualla, Jonathan Ho, Andrea Tagliasacchi, and Mohammad Norouzi.
\newblock Novel view synthesis with diffusion models.
\newblock {\em arXiv preprint arXiv:2210.04628}, 2022.

\bibitem{wu2023multiview}
Chao-Yuan Wu, Justin Johnson, Jitendra Malik, Christoph Feichtenhofer, and Georgia Gkioxari.
\newblock Multiview compressive coding for 3{D} reconstruction.
\newblock {\em arXiv:2301.08247}, 2023.

\bibitem{wu2023reconfusion}
Rundi Wu, Ben Mildenhall, Philipp Henzler, Keunhong Park, Ruiqi Gao, Daniel Watson, Pratul~P. Srinivasan, Dor Verbin, Jonathan~T. Barron, Ben Poole, and Aleksander Holynski.
\newblock Reconfusion: 3d reconstruction with diffusion priors.
\newblock {\em arXiv}, 2023.

\bibitem{xu2019understanding}
Jingjing Xu, Xu~Sun, Zhiyuan Zhang, Guangxiang Zhao, and Junyang Lin.
\newblock Understanding and improving layer normalization.
\newblock In {\em Advances in Neural Information Processing Systems}. NeurIPS, 2019.

\bibitem{xu2023dmv3d}
Yinghao Xu, Hao Tan, Fujun Luan, Sai Bi, Peng Wang, Jiahao Li, Zifan Shi, Kalyan Sunkavalli, Gordon Wetzstein, Zexiang Xu, et~al.
\newblock Dmv3d: Denoising multi-view diffusion using 3d large reconstruction model.
\newblock {\em arXiv preprint arXiv:2311.09217}, 2023.

\bibitem{tilted2023}
Brent Yi, Weijia Zeng, Sam Buchanan, and Yi~Ma.
\newblock Canonical factors for hybrid neural fields.
\newblock In {\em International Conference on Computer Vision (ICCV)}, 2023.

\bibitem{yu2021plenoxels}
Alex Yu, Sara Fridovich-Keil, Matthew Tancik, Qinhong Chen, Benjamin Recht, and Angjoo Kanazawa.
\newblock Plenoxels: Radiance fields without neural networks.
\newblock {\em arXiv preprint arXiv:2112.05131}, 2021.

\bibitem{yu2021pixelnerf}
Alex Yu, Vickie Ye, Matthew Tancik, and Angjoo Kanazawa.
\newblock pixelnerf: Neural radiance fields from one or few images.
\newblock In {\em Proceedings of the IEEE/CVF Conference on Computer Vision and Pattern Recognition}, pages 4578--4587, 2021.

\bibitem{yu2023mvimgnet}
Xianggang Yu, Mutian Xu, Yidan Zhang, Haolin Liu, Chongjie Ye, Yushuang Wu, Zizheng Yan, Chenming Zhu, Zhangyang Xiong, Tianyou Liang, et~al.
\newblock Mvimgnet: A large-scale dataset of multi-view images.
\newblock In {\em Proceedings of the IEEE/CVF Conference on Computer Vision and Pattern Recognition}, pages 9150--9161, 2023.

\bibitem{zhang2018unreasonable}
Richard Zhang, Phillip Isola, Alexei~A Efros, Eli Shechtman, and Oliver Wang.
\newblock The unreasonable effectiveness of deep features as a perceptual metric.
\newblock In {\em Proceedings of the IEEE conference on computer vision and pattern recognition}, pages 586--595, 2018.

\end{thebibliography}

\end{document}